# Quran Intelligent Ontology Construction Approach Using Association Rules Mining

Fouzi Harrag,
Computer Sciences Department,
College of Sciences,
Ferhat Abbas University,
Setif, 19000, Algeria,
fouzi.harrag@univ-setif.dz,

Abdullah Al-Nasser[2], Abdullah Al-Musnad[3], Rayan Al-Shaya[4], Abdulmalik Salman Al-Salman[5]
Computer Sciences Department,
College of computer & Information Sciences,
King Saud University,
Riyadh, 11543, Saudi Arabia,
[2]a.alnassr@hotmail.com, [3]amm4649@gmail.com,
[4]rayan.a.sh.91@gmail.com, [5]salman@ksu.edu.sa

*Abstract*— Ontology can be seen as a formal representation of knowledge. They have been investigated in many artificial intelligence studies including semantic web, software engineering, and information retrieval. The aim of ontology is to develop knowledge representations that can be shared and reused. This research project is concerned with the use of association rules to extract the Quran ontology. The manual acquisition of ontologies from Quran verses can be very costly; therefore, we need an intelligent system for Quran ontology construction using pattern-based schemes and associations rules to discover Quran concepts and semantics relations from Quran verses. Our system is based on the combination of statistics and linguistics methods to extract concepts and conceptual relations from Quran. In particular, a linguistic pattern-based approach is exploited to extract specific concepts from the Quran, while the conceptual relations are found based on association rules technique. The Quran ontology will offer a new and powerful representation of Quran knowledge, and the association rules will help to represent the relations between all classes of connected concepts in the Quran ontology.

## I. INTRODUCTION

In recent years the notion of ontology has emerged and become common for researchers and scientists in the world. Web Ontologies can be ranged from large taxonomies categorizing Web sites (such as on Yahoo, Ask and Aol) to categorizations of products for sale and their features (such as on Amazon.com and ebay.com). Ontology provides a solution to capture information about concepts and relations between those concepts in the same domain [17]. Guber et al, [6] defined ontology as "A formal explicit specification of a shared conceptualization." Ontology can be represented in a declarative form by using web semantic languages such as RDF, OWL or XML. There are many potential benefits of using ontology in representing and processing knowledge, including the separation of domain knowledge from application knowledge, sharing of common knowledge of subjects among human and computers, and the reuse of domain knowledge for a variety of applications.

This paper aims to create an ontology extraction tool for Holy Quran using the association rule algorithm. This involves investigating the most proper methods for dealing with the Quran's text collection to extract concepts and semantic relations between them. Using this Quran ontology, various text mining tasks including information extraction, text categorization, concept linkage, and discovery of associations and patterns can be tested and evaluated. The paper considers investigating the Quran text as a pilot benchmark before focusing later on the entire Islamic studies collection to build a fully functional ontology for Islam. The ontology will provide a powerful representation of Quran knowledge, with the rule schemas giving a more expressive representation of Quran relations in term of rules.

## II. LITERATURE REVIEW

Ontologies provide a framework for handling structured information and extracting conclusions from the structured information [4]. The term ontology is used in information and knowledge representation systems to denote a knowledge model, which represents a particular domain of interest [14]. A body of formally represented knowledge is based on a conceptualization: the objects, concepts, and other entities that are assumed to exist in some area of interest and the relationships that exist among them [8]. The goal of a domain ontology is to reduce the conceptual and terminological confusion among the members of a virtual community of users that need to share electronic documents and various kinds of information. Many ontologies learnt from text applications have been published and presented in various domains. Trappey et al. [15] presented a novel ontology schema based on a hierarchical clustering approach for knowledge document self-organization, particularly for patent analysis. Sharif et al., [14] tried to present a simple ontology of folksonomy to show how different elements act in such a dynamic space, and how implicit relations emerge from implicit complex networks within the folksonomies. Marinica and Guillet [10] set out to improve post-processing of association rules by a better integration of user (decision maker) goals and knowledge. Tatsiopoulos and Boutsinas [16] presented a new ontology mapping technique, which given two input ontologies, is able to map concepts in one ontology onto those in the other without any user intervention. Lau et al., [9] proposed a novel concept map generation mechanism, which is underpinned by a fuzzy domain ontology extraction algorithm.

## A. Arabic Ontologies

Elkateb et al., [5] introduced the Arabic WordNet project that was generated using the same methodology used to build EuroWordNet. The Suggested Upper Merged Ontology (SUMO)1 is used to map between different languages. Belkredim and Meziane, [2] proposed DEAR-ONTO, a derivational Arabic ontology that represents the Arabic language. They classified verbs as sound verbs and weak ones, and using measures and derivation they built an ontology for the Arabic language. Saad et al., [12] presented a general methodology to extract information from the Islamic Knowledge sources to build an ontology for a given domain. Their approach based on combination of natural language processing techniques, Information Extraction and Text Mining techniques. In the Quran corpus [3], a knowledge representation is used to define the key concepts in the Quran, and to show the relationships between these concepts using predicate logic. Named entities in verses, such as the names of historic people and places mentioned in the Quran, are linked to concepts in the ontology as part of named entity tagging. Harrag et al., [7] investigated the use of association rules to identify frequent itemsets over concepts that are related to Islamic jurisprudence (Fiqh). The semantic structure of the Sahîh Al-Bukhârî as a knowledge source was exploited to extract specific domain ontology, while the conceptual relations embedded in this knowledge source were modeled based on the notion of association rules.

## III. APPROACH

In this section we will provide a depth look on our approach for constructing the Quran Ontology, so we will describe the Quran corpus and we will define the steps to build an ontolgy for this corpus.

### A. Quran Corpus

The Quran (Arabic: القرآن literally meaning "the recitation", also romanised Qur'an or Koran) is the central religious text of Islam. Muslims believe the Qur'an to be the book of divine guidance and direction for mankind. They also consider the text in its original Arabic, to be the literal word of Allah[1] revealed by the angel Jibreel (Gabriel) to Muhammad (صلى الله عليه و سلم) both in word and in meaning over a period of twenty-three years. It is collected between the two covers of the mushaaf, was narrated in mutawaatir chains, and is a challenge to humankind. Muslims consider the Quran to be the only book that has been protected by God from distortion or corruption. Quranic chapters are called suras and verses are called ayahs.

As can be seen from many studies on the Quran, general statistics about the text of the Quran are as follows:

- The number of chapters that start with Basmalah, is 113.
- Only one chapter does not start with Basmalah and that is chapter At-Tawbah.
- There are 30 parts (juz') in the Quran. A part is two sections (Hizb).
- There are 60 sections in the Quran. A section is divided into 4 quarters.
- There are 240 quarters in the Quran.
- Number of letters is 330,709
- Number of words without repetition is 14,870
- Number of words in the entire Quran is 77,797
- Number of verses is 6,236

We could also deduce that the average word length in the Quran is 330,709 ÷ 77,797 = 4.25. Figure 1 shows the

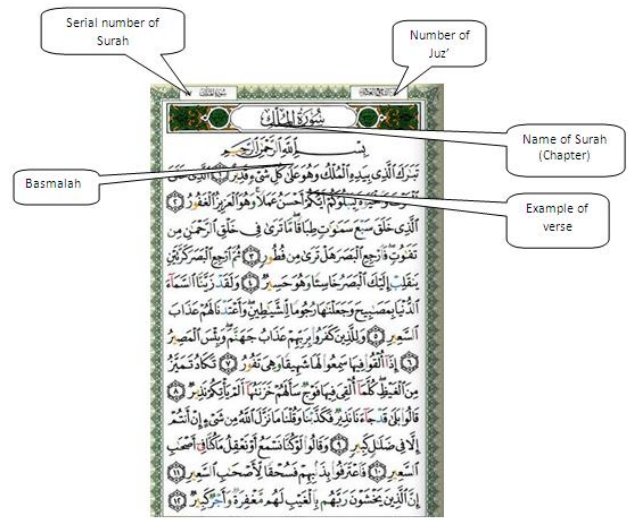

Fig. 1. Structure of Surat Al-Mulk.

### B. Approach description

In this paper, we will develop techniques to mine an ontology from Quran texts. The ontology will constitute a specific vocabulary used to describe a particular model of the Islamic world as shown in figure 2. The main steps are the following:

1. A basic framework for the Quran knowledge ontology will be developed based on existing knowledge representations of the Semantic Web.
2. Ontological knowledge representation enables domain experts to define knowledge in a consistent way using a standard format (such as XML, RDF, or OWL).
3. Based on the linguistic expressions of certain Quran text, we will identify and develop an association rules based algorithm for extracting different Quran ontological layers from the Quran texts.
4. The different components of text will serve as input features and the different ontological layers based on the Quran ontology as outputs of the system.

---

[1] SUMO: Suggested Upper Merged Ontology, http://www.ontologyportal.org/

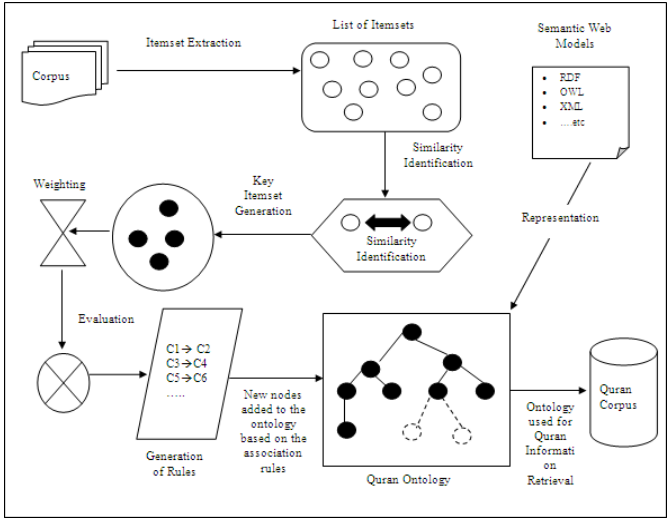

Fig. 2. Main steps of our approach for Quran ontology construction

The main phases of our approach are explained below.

*1) Phase I:*
A basic framework for the Quran knowledge ontology of the Holy Quran texts will be developed based on existing knowledge representations of the Semantic Web. Ontological knowledge representation enables domain experts to define knowledge in a consistent way using a standard format (such as XML, Resource Description Framework (RDF), or Web Ontology Language (OWL)).

*2) Phase II:*
Based on the linguistic expressions of certain Quran texts, we will identify and develop an association rules based algorithm, to extract different Quran ontological layers from the Holy-Quran texts. In our work, a number of steps, as detailed below, are needed to extract the proposed Quran ontology with similarity degrees between items. The mining process can generate more relevant association rules, based on embedded knowledge within the Quran texts.

a) *Itemset extraction:* The first step is itemset extraction, which identifies items in the Quran corpus and generates itemsets of length one.

b) *Similarity identification:* Similarity degree values between items are provided by the Quran ontology, which specifies the semantics of the Quran corpus content. Such ontology can be created by Quran experts or reused from existing ones available on the Semantic Web. Once the Quran ontology has been defined, a semantic similarity association is detected. These pairs of items will contribute to the composition of association rules.

c) *Generation of key itemsets:* This step generates key itemsets from the items identified in the first and second steps.

d) *Weighting the key itemsets:* After the generation of key itemsets, a weighting, which consists of the count of the number of occurrences of a corresponding itemset in the Quran corpus, will be applied to each itemset.

e) *Evaluation using the Apriori algorithm:* Apriori (Agrawal and Srikant, 1994) is the best-known algorithm to mine association rules. It uses a breadth-first search strategy to counting the support of itemsets and uses a candidate generation function which exploits the downward closure property of support. In this step we evaluate the support and confidence of each selected itemset. If the support of a key itemset is greater than or equal to a predefined threshold "M-S" (for Minimum-Support), the corresponding itemset is added to the set of frequent itemsets, which will be used to generate association rules.

f) *Generating rules:* In the last step, generalization possibilities are sought for each frequent itemset. If the confidence of a rule is greater than or equal to a predefined threshold "M-C" (for Minimum-Confidence), the rule is considered valid. Association rule generation is usually split up into two separate steps:

- First, minimum support is applied to find all frequent itemsets in a database.
- Second, these frequent itemsets and the minimum confidence constraint are used to form rules.

*3) Phase III:*
The different components of the text will serve as input features for this phase, while the different ontological layers based on the Quran ontology will be output by the system.

*4) Phase IV:*
An ontology-based approach is used to improve Quran information retrieval (IR). The IR system is based on a domain knowledge representation schema in the form of domain ontology.

IV. ONTOLOGY CONSTRUCTION

Our work is based on building ontology for the holy Quran Chapters (*Surah*) related to the stories of the prophets. Our ontology represents the main concepts of the Verses (*Al-Ayat*) as semantic relationships. By using this ontology, we can search for any word/phrase and find the related Chapter (*Surat*) and verses (*Ayat*) based on some relationships that will be discussed in the next sections.

Our work consists of building ontology from the set of concepts extracted out of the texts of Quran and the set of the relations between these concepts. We follow the same steps of the methodology described in [11]. Table 1 show an example of some concepts and relations from our ontology.

**Table.1** Concepts and relations from Quran ontology.

| Concept | Relation | Example |
|---------|----------|---------|
| Words or (concepts) that exist in the text of the Verses: Prophet, Oumah, Book, Islam, Mohamed, Mos ….etc . | Many relations like: Part of, Synonym, kind of,...,etc . | أولي العزم **is a part of** الرسل .<br><br>القرآن **is a kind of** الكتب<br><br>اليهود **is Synonym of** بنو إسرئيل |

Figure 3 show an overview of our methodology for the construction of the Quran ontology. The main steps are the following:

*A. Determining the domain and the scope of our ontology*

From the whole Islamic knowledge domain, we focus our attention on the *Quran* text due to its importance in the Muslims life. In our work, we opted for the use of a collection of Chapters (*Surah*) related to the prophets' stories in the Quran as a domain for our ontology.

*B. Extracting the important terms in the ontology*

Before starting the process of our ontology building, we have to list all the important terms in the texts to facilitate the next operation of concepts extraction. Arabic Natural language processing (NLP) tools are used for this goal. The *KP-Miner*[2] System is considered for this preprocessing step. This system applies an operation of tagging to all the words of the texts and after that it starts the operation of key-phrases extraction. These operations will allow us to reduce the domain of our ontology by focusing only on verbs, nouns and multi-words.

*C. Defining the concepts*

This step is dedicated to the operation of concepts extraction. The inputs of this step are the outputs of the previous one i.e. the important terms extracted by *KP-Miner*. In this step we apply the stemming operation. Stemming is concerned with the transformation of all derivatives words to their single common stem or canonical form. This process is very useful in terms of reducing and compressing the indexing structure, and it taking advantage of the semantic/conceptual relationships between the different forms of the same root. A modified version of *Shereen Khoja*[3] stemmer has been used to complete this step.

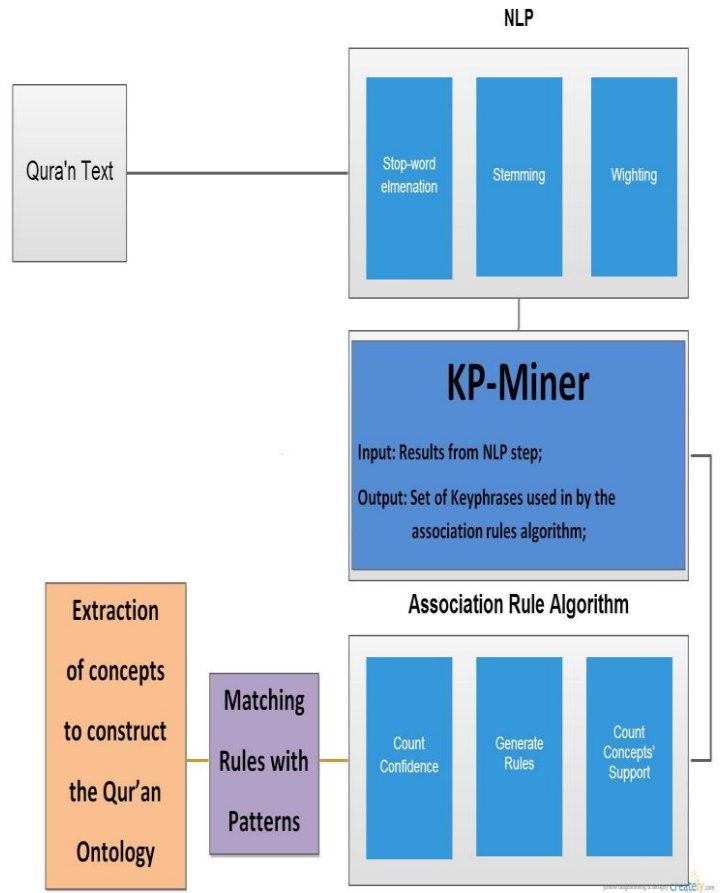

Fig. 3. Extracting the important terms in the ontology.

*D. Defining the relationships between concepts*

Once we have defined the concepts, we must define the relationship between them. This step is the most important one in the process of building for our ontology. This importance is due to the idea that ontology is based on the representation of the extracted concepts with their relationships. Actually, there are many algorithms and techniques to extract relations between concepts, but the problem is that some of them don't fit to the Arabic language. For this purpose, we opt for the use of association rules method to extract relations, this choice is justified by the idea that associations rules don't depend on any linguistic consideration. Two important points must be mentioned before starting the discussion of our relations extraction method. In the association rules method, the rule (A → B) represents the relationship between the two concepts A and B. The concept B will be considered in a higher conceptual level than the concept A. Let's take for example the rule: نوح → الأنبياء . The concept الأنبياء is located in a higher conceptual level than the concept نوح. The sub-concept نوح is a kind of the super-concept الأنبياء. To apply the association rules method, we represent all the verses of our corpus as instance of transaction as it is mentioned in table 2.

---

[2] The *KP-Miner* System: http://www.claes.sci.eg/coe_wm/kpminer/. Last Visited: September 15, 2013.

[3] *Shereen Khoja* stemmer: http://zeus.cs.pacificu.edu/shereen/research.htm. Last Visited: September 15, 2013.

**Table.2** Representation of Quran verses as instance of transaction.

| ID | Verses |
|---|---|
| 1 | لقد **أرسلنا نوحا** إلى **قومه** فقال يا قوم اعبدوا الله ما لكم من إله غيره إني أخاف عليكم عذاب يوم عظيم |
| 2 | **وإلى عاد** أخاهم **هودا** قال يا قوم اعبدوا الله ما لكم من إله غيره أفلا تتقون |
| 3 | **وإلى ثمود** أخاهم **صالحا** قال يا قوم اعبدوا الله ما لكم من إله غيره قد جاءتكم بينة من ربكم هذه ناقة الله لكم آية فذروها تأكل في أرض الله ولا تمسوها بسوء فيأخذكم عذاب أليم |
| …… | …… |

The Quran ontology represents the semantic relations between the different concepts extracted from the corpus of *Stories of the Prophets* in the Qur'an. Multiple instances of these relations are of type "part of", "kind of" and "synonym of". Our approach is based on assigning a set of tags to each type of relation; these tags will be used to detect the type of the relations and the concepts concerned by this relation.

If we consider the different type of relations as: $R_i = \{R_1, R_2, R_3, R_4, \ldots, R_n\} / i=1..n$, then for each relation $R_k$, we define a set of **tags** $T_j = \{T_1, T_2, T_3, \ldots, T_m\} / j=1..m$. if any tag $T_j$ from $T$ is found in the text of the hadith, this means the existence of a relation of type $R_i$:

$$T_k \in M; \quad M = T_k AB \rightarrow \exists R_k / A \stackrel{R_k}{\Rightarrow} B \qquad (1)$$

The method of relation's extraction can be summarized as follows:

- **Step-1**: for each relation, find all tags that co-occurs with the higher concept (B) of the relation R.
- **Step-2**: apply the Apriori algorithm by finding all the itemsets of concepts with (*K=2*) and *min_support=1*, then find all possible association rule between concepts in each *verses*.
- **Step-3**: Determine the position of the tag word in the relation and then delete all the rules that don't satisfy the condition defined in step 1.
- **Step-4**: Find the confidence for each rule and accept only the rule that satisfy (confidence >= min_ confidence). The confidence of the rule $A \stackrel{x}{\rightarrow} B$ is found by counting the number of occurrence of the itemsets {A, B} in all *verses*.

V. EXPERIMENTAL RESULTS

In order to measure the performance of our system, a collection of Chapters (*Surahs*) from the holy Qur'an is used. We used a specific corpus characterized by the specialization of its domain. It includes 12 *Surahs* related to the stories of prophets in the Qur'an as depicted in Table 3. Our dataset contains 1407 verses (Ayah). The number of words collected from these *Surahs* after the preprocessing step is 16153 words. Table 3 represents the number of Ayahs and the number of words for each Surah in this corpus.

**Table 3.** Number of Ayah and word per Surah for our Qur'an Corpus

| Surah name | | Number of Ayah | Number of word |
|---|---|---|---|
| Al-Aaraf | الأعراف | 206 | 3344 |
| Al-Anbiaa | الأنبياء | 112 | 1174 |
| Al-Sajda | السجدة | 30 | 374 |
| Al-Shuaara | الشعراء | 227 | 1322 |
| Al-Shuraa | الشورى | 53 | 860 |
| Al-Safat | الصافات | 182 | 865 |
| Al-Ankabut | العنكبوت | 69 | 982 |
| Al-Mominun | المؤمنون | 118 | 1051 |
| Al-Naml | النمل | 93 | 1165 |
| Ghafir | غافر | 85 | 1228 |
| Hud | هود | 123 | 1947 |
| Yunes | يونس | 109 | 1841 |
| **Total** | | **1407** | **16153** |

*A. Feature selection using Support filters*

The Apriori algorithm will be applied on the outcomes of the NLP process. We will compute the support of each concept that comes from the NLP step. The Support of concept $C_i$ is the number of transaction containing the Concept $C_i$ divided by the total number of transaction.

$$\text{Sup}_{Ci} = \frac{N_i}{N} \qquad (2)$$

Each *Ayah* will be considered as a single transaction. After computing the support for all concepts, we will retain only the concepts that take their values in a predefined range. The

average of concept's support will be used to determine this range. We will consider the following formulas:

$$\text{ave} = \frac{\sum_0^n s_i}{n} \qquad (3)$$

$$\text{max} = \text{ave} + \frac{\text{ave}}{2} \qquad (4)$$

$$\text{min} = \text{aev} - \frac{\text{ave}}{2} \qquad (5)$$

Table 4 shows an example of support's results for the ten top concepts of *Surat "Al-Mumtahana"*, after applying the NLP process.

**Table.4** Support's results for the top ten concepts of *Surat "Al-Mumtahana"*

| Term | Support |
|---|---|
| بينكم | 0.307 |
| الله | 0.769 |
| يا | 0.307 |
| جاءكم | 0.153 |
| تؤمنوا | 0.153 |
| انفقوا | 0.153 |
| هن | 0.153 |
| ربنا | 0.153 |
| مؤمن | 0.230 |
| كفروا | 0.153 |

By using formulas 1, 2 and 3 on these results, the average of all concepts' supports was equal to 0.18518, the max was equal to 0.27778 and the min was equal to 0.09259. These values are then used to filter the results and to ignore the very rare and the very frequents concepts.

*B. Confidence and rules generation*

The confidence for the rule A, B → C will be computed by using the formula 6:

$$\text{Con} = \frac{\text{Sup}(A \cup B \cup C)}{\text{Sup}(A \cup B)} \qquad (6)$$

After the calculation of supports for all concepts, the next step is to generate the association rules from these concepts. Once the rules are generated, another elimination step will be applied to remove those rules out of the confidence range. Max and min filters will be computed the same way as the max and min of the support range.

Table 5 shows an example of the top ten rules generated from the concepts of *Surat "Al-Mumtahana"*:

**Table.5** Confidence's results for the top ten rules extracted from *Surat "Al-Mumtahana"*

| Rule | Confidence |
|---|---|
| انفقوا → هن , جاءكم | 2 |
| هن → تؤمنوا , جاءكم | 3 |
| تؤمنوا → انفقوا , جاءكم | 1.5 |
| مؤمن → هن , ربنا | 2.5 |
| كفروا → هن , ربنا | 1.5 |
| مؤمن → تؤمنوا , جاءكم | 1.5 |
| مؤمن → جاءكم , كفروا | 3 |
| كفروا → ربنا , جاءكم | 3 |
| تؤمنوا → ربنا , انفقوا | 1.5 |
| هن → تؤمنوا , كفروا | 2 |

*C. Concepts extraction using triggers:*

After the generation of association rules using the *Apriori* algorithm, the result consists of a huge number of rules; some of them have no meaning at all. Some others have mining but not related to the domain of our study (prophetic stories). To solve this problem we will specify certain words to pick up the rules that match our interest. These words are called triggers.

Table 6 shows an example from the list of triggers considered in our system.

**Table.6** Example of triggers used in the extraction task.

| Trigger | Meaning |
|---|---|
| وهبنا | Trigger specified for the relation (prophet نبي – progeny ذرية). |
| أرسلنا | Trigger specified for the relation (prophet نبي – nation قوم). |
| آتينا | Trigger specified for the relation (prophet نبي - holy book كتاب مقدس ). |

These triggers can't give the full solution by themselves, so we will combine them with certain terms to form a set of patterns that we will be used to valid the strong rules and to extract possible concepts related to the scope of our ontology. Table 7 shows the list of patterns formed from the previous triggers.

**Table.7** Example of patterns used in the extraction task

| Trigger | Pattern | Meaning |
|---|---|---|
| وهبنا | نبي + وهبنا --> نبي | Trigger وهبنا often come with a name of prophet to show the relation with his progeny. |
| أرسلنا | نبي + أرسلنا --> قوم | Trigger أرسلنا sometimes come with a name of prophet to show the relation with his nation. |
|  | قوم + أرسلنا --> عذاب | Trigger أرسلنا other times come with a name of nation to show the kind of torment |
| آتينا | نبي + آتينا --> كتاب | Trigger آتينا often come with an name of prophet to show the relation with the holy book |

These patterns are considered as an additional filter for the set of generated rules. Only the rule that match these patterns will be accepted, the others will be rejected. Table 8 shows an example of the application of patterns to the set of the generated rules from *Surat "Al-A'raf"*.

**Table.8** Example of application of patterns to the rules of *Surat "Al-A'raf"*.

| Rule | Matched pattern | Accepted/Rejected |
|---|---|---|
| رجل + بعثنا --> كافر | -- | Rejected |
| هدية + سماء --> نفس | -- | Rejected |
| آدم + آمنوا --> خالدون | نبي + آمنوا --> صفة حسنة | Accepted |
| آدم + آمنوا --> عدو | نبي + آمنوا --> صفة حسنة | Accepted |
| ربكم + بالحق --> أنزلنا | -- | Rejected |
| نفس + عمل --> ملك | -- | Rejected |
| رسل + أرسلنا --> بالحق | -- | Rejected |
| آدم + أرسلنا --> الطيبات | نبي + أرسلنا --> قوم | Accepted |
| آدم + أرسلنا --> أمة | نبي + أرسلنا --> قوم | Accepted |
| آدم + بعثنا --> سلطانا | نبي + بعثنا --> قوم | Accepted |

As we see in this example, only five out of ten rules are accepted. That will be very efficient to reduce the huge number of rule coming from association rule algorithm. Note that we only take the first two concepts into consideration.

An overview of the obtained results is given in table 9. We can distinguish the clear difference between the number of rules generated by the Apriori algorithm and the number of rules after applying the patterns.

**Table.9** Overview of the results obtained on our Dataset.

| Surat | KP-Miner Outputs | # of Extracted Rules | # of Rules after applaying Patterns |
|---|---|---|---|
| الأعراف | 149 | ≈500000 | 148 |
| الأنبياء | 111 | 444752 | 220 |
| السجدة | 53 | 159606 | 0 |
| الشعراء | 91 | 41490 | 360 |
| الشورى | 80 | 172994 | 0 |
| الصافات | 69 | 58461 | 408 |
| العنكبوت | 97 | 408600 | 0 |
| المؤمنون | 93 | ≈500000 | 92 |
| النمل | 113 | 328428 | 672 |
| غافر | 109 | ≈500000 | 864 |
| هود | 120 | 217351 | 238 |
| يونس | 130 | ≈500000 | 129 |

VI. CONCLUSION

In this paper, we were interested on the use of association rules method in the process of concepts extraction in order to build an ontology for the Quran's corpus. The approach was based on using association rules to identify strong rules over the concepts extracted from the Quran texts by generating a set of accepted rules using the Apriori algorithm. The set of concepts extracted from the texts of the whole Quran and their conceptual relations was modeled based on the notion of association rules to construct this ontology. The generated ontology will be used as new representation tool for the Quran.

*Acknowledgment*

This work was supported by the Research Center of College of Computer and Information Sciences, King Saud University under the project code (RC121225). The author is grateful for this support.